\documentclass[a4paper, 10pt, conference]{ieeeconf}      % Use this line for a4 paper

\IEEEoverridecommandlockouts                              % This command is only needed if 
\usepackage{amsmath} % assumes amsmath package installed
\usepackage{graphicx}
\usepackage{comment}
\usepackage{subfig} % qilu 01/29/2024 make it possible to include more than one captioned figure/table in a single float
\usepackage{xcolor}
\usepackage{algorithm}
\usepackage{cite}
\usepackage{url} 
\usepackage[noend]{algpseudocode}

\title{\LARGE \bf
A Visitation Grid for Complete Coverage Foraging in Robot Swarms }

\author{Arturo Gonzalez, Yifeng Gao, Li Zhang, and Qi Lu 
\\ Department of Computer Science 
\\ The University of Texas Rio Grande Valley, Texas, United States 
\\ \{arturo.gonzalez11, yifeng.gao, li.zhang, qi.lu\}@utrgv.edu
}

\begin{document}

\maketitle
\thispagestyle{empty}
\pagestyle{empty}

%%%%%%%%%%%%%%%%%%%%%%%%%%%%%%%%%%%%%%%%%%%%%%%%%%%%%%%%%%%%%%%%%%%%%%%%%%%%%%%%
\begin{abstract}
The complete collection of sparse resources in large, unknown environments remains a challenging problem for autonomous robot swarms. Previous studies have shown that a substantial portion of total mission time is consumed during the final stage of collection, where only a small fraction of randomly scattered resources remain. Consequently, many existing swarm foraging algorithms (search and collection) focus on collecting most resources within a limited time window, rather than improving end-stage efficiency for collecting all resources. We propose a grid-based stochastic foraging strategy that explicitly reduces redundant visits and accelerates late-stage collection. The unknown search area is partitioned into a grid map, which is maintained by a lightweight central server. To maintain scalability, both robots and the server operate within limited memory and computational constraints. Robots execute decentralized search behaviors while exploring assigned regions and communicate summarized trajectory information to the server upon returning to the center only. The server updates the grid-level visitation counts based on robot-reported locations, producing a global estimate of the exploration density. For each new foraging trip, a robot selects its next search area from a local $3\times 3$ neighborhood of grids probabilistically with the lowest visitation count, thus biasing exploration toward under-visited regions while maintaining stochasticity. Extensive simulation experiments demonstrate that the proposed strategy consistently outperforms the canonical centrally placed baseline foraging algorithm (CPFA). Compared to CPFA, the proposed method reduces the total collection time by up to $33\%$ and improves collection efficiency by more than $48\%$ during the final stage of the mission. These results indicate that the proposed strategy is robust, flexible, and scalable for near-complete and complete resource collection in robot swarms and can serve as a general enhancement for stochastic swarm foraging methods under limited onboard resources.
\end{abstract}

%\begin{IEEEkeywords}
%Swarm Robotics, Intrusion Detection, Foraging Robots, Pheromone Trails, Security in Robotics
%\end{IEEEkeywords}

\section{Introduction}
\label{Introduction}

Swarm robotics investigates the coordination of numerous simple, autonomous robots to accomplish complex tasks efficiently. Contemporary research in this field encompasses a range of collective behaviors, including self-organization~\cite{self-org1,self-org2}, task allocation~\cite{acbba2020}, aggregation~\cite{cue-based1,cue-based2}, object sorting~\cite{sorting0,sorting1}, and foraging~\cite{Lu2018DynamicDepots, QualitySensitiveForaging2018,ForagingDRL2020,2021RobotChain,kaminka2025heterogeneous}. 

%The foraging task is an abstraction of a variety of critical real-world applications, including search and rescue, environmental monitoring, planetary exploration, and disaster response.

The foraging task is originally from natural systems like ant colonies, termites, schools of fish, and flocks of birds. It involves searching for resources (e.g., seeds) collectively in a large, unknown arena and delivering them to a specified location (e.g., nests). We developed collaborative foraging algorithms for robot swarms (a group of robots) to search for resources as much as possible in a certain time window. In~\cite{BeyondPherom2015,2016IROSMPFA, Lu2018DynamicDepots, MatthewDDSA2016}, robot swarms utilize virtual pheromone trails to direct their search toward clustered resources - individual robots are more likely to remember or communicate the locations of resources found in dense clusters. Existing work demonstrates that resource intake rates are efficient when the density of resources is high. Rates tend to drop sharply when only a small, sparsely distributed fraction of residual resources remains. In~\cite{CompleteCollection2015}, the experimental results indicate that swarms foraging with the Central Place Foraging Algorithm (CPFA) spend between $63\%$ and $75\%$ of their time collecting the last $12\%$ of the total resources.

% In this work, we proposed a Grid-Based Complete Foraging Algorithm (GCFA) to improve the foraging performance of robot swarms in collecting scarce resources. We assume that robots can only record a limited number of visited locations at a certain frequency. They share visited recorded locations with the server in the central collection zone when they return to the collection zone. In this case, the robot swarm is scalable when communication is limited. The server organizes the shared information in a grid-based map. Robots go to the least visited regions, probabilistically based on the information in the grid, and search for remaining resources in the regions. 

A complete collection of key resources is crucial in various real-world applications, including search and rescue, humanitarian demining, and the detection of chemical leaks and intruders. In this work, we propose a visitation grid-guided stochastic foraging strategy that explicitly steers robots away from already heavily explored areas to the least-visited areas. The basic idea is straightforward: robots periodically record visited positions into a bounded local queue, upload this data to a central server upon returning, and the server accumulates per-cell visit counts in the grid map. Each robot departing in uninformed search mode is then probabilistically directed toward the least-visited neighborhood of the arena, reducing redundant revisits without requiring direct robot-to-robot communication or unbounded onboard storage.

We name this strategy the Grid-Based Complete Foraging Algorithm (GCFA). The central server partitions the search arena into a finite grid. The server stores per-cell visitation counters ($O(n^2)$ memory) rather than every single reported visitation coordinate in the arena. Therefore, the selection of the next visitation region runs in $O(n^2)$ time. The design is discussed in Section~\ref{method} in detail.

The paper proceeds to detail previous work on swarm robotics foraging algorithms (Section \ref{related_work}), the central-placed foraging algorithm (Section \ref{background}), the methodology for the proposed GCFA (Section \ref{method}), our experimental setup (Section \ref{experiment}), and the evaluation of our experiments (Section \ref{results}). We conclude with a summary of our contributions and directions for future work (Section \ref{conclusion}).

\section{Related work}
\label{related_work}

There are also many works related to foraging robot swarms. The work in~\cite{BeyondPherom2015} first introduced the centrally-placed CPFA foraging algorithm, which uses a single central collection zone in the environment. It is a stochastic search algorithm in which robots search for resources randomly and then share information about the discovered resources. Robots are recruited to the locations where robots discovered resources by virtual pheromone waypoints. Some locations may be visited multiple times, and others may not be. The relationship between resource quality and distances from the pheromone trail is described in~\cite{Sophisticated2020}, which presents a collective foraging system based on virtual pheromones. This study demonstrates the sufficiency of simple individual robot rules to generate sophisticated collective emerging foraging behavior. 

Beyond pheromone-based approaches, frontier-based exploration strategies have been extensively studied for systematic area coverage. The work in~\cite{frontierbased1997} introduces frontier-based exploration where robots identify boundaries between explored and unexplored regions, ensuring comprehensive coverage. This has been extended to multi-robot coordination in~\cite{coordinatedexploration2005}, where robots select distinct frontiers to minimize redundant exploration. Alternative negative feedback mechanisms have also been proposed, such as the stigmergic approach in~\cite{evaporatingtraces1999}, which implements virtual repellents that discourage robots from revisiting recently explored areas. These approaches share conceptual similarities with our visit-count guided strategy in maintaining spatial memory to guide exploration, though our method leverages centralized grid-based aggregation rather than distributed frontier selection or stigmergic markers.

In our previous work~\cite{2016IROSMPFA, Lu2018DynamicDepots}, we proposed a multiple-placed foraging algorithm, which further improves the foraging performance. Multiple collection zones or multiple dynamic depots are deployed in the environment. Robots deliver resources to their closest depots. The depots move to the centroid of the discovered locations where the robots found resources. Therefore, multiple depots reduce the congestion around the central collection zone in the CPFA, and robots travel shorter distances to deliver resources.  

The work in~\cite{MatthewDDSA2016} presents a deterministic spiral search algorithm. Robots search for resources in pre-planned multiple interlocking spiral paths. Robots start the search from the center and search for resources on the spiral paths. The locations close to the center are visited early, and the locations far from the center are visited later. Finally, all locations will be visited. This algorithm outperforms the CPFA algorithm when resources are close to the center. However, if the search arena is large, it takes a much longer time to find them. Additionally, the deterministic search strategy is not flexible in a real-world environment. It can not visit locations accurately since there is always noise in the localization. It also does not work efficiently in an environment with obstacles~\cite{Lu2018DynamicDepots}.

However, robots collect a certain percentage of resources in all these related works. They do not complete the collection of all the resources. In~\cite{CompleteCollection2015}, the experimental results indicate that the foraging performance decreases dramatically when there are fewer than $12\%$ resources in the environment. The time to collect the first $88\%$ of the resources is approximately half of the time to collect the last $12\%$ of the resources. In this work, we proposed the visit-count guided GCFA algorithm to improve the foraging performance for the complete collection of the last $12\%$ of resources. 

Beyond biologically inspired search, the coordination of robot swarms in unpredictable settings has been influenced by the $m$-vehicle Dynamic Traveling Repairman Problem ($m$-DTRP)~\cite{Frazzoli2004, Pavone2011}. This framework addresses decentralized motion coordination to minimize the expected waiting time for stochastically generated targets~\cite{Frazzoli2004, Pavone2011}. Advanced control policies within this domain, such as the Receding Horizon (RH) Median/TSP and Divide \& Conquer (DC) strategies, demonstrate how swarms can remain adaptive and scalable by partitioning the environment into equitable subregions~\cite{Frazzoli2004, Pavone2011}.

While $m$-DTRP strategies are optimized for managing continuous streams of new demands, our GCFA focuses on the unique challenge of the "endgame" phase in finite resource collection. Unlike the partitioning methods in the routing literature that primarily use Voronoi cells to balance loads~\cite{Frazzoli2004, Pavone2011}, our visit-count strategy provides a lightweight spatial memory specifically designed to reduce redundant revisits in resource-sparse environments.
%Beyond pheromone-based approaches, frontier-based exploration strategies have been extensively studied for systematic area coverage. The work in~\cite{Jackson2007} introduces frontier-based exploration where robots identify boundaries between explored and unexplored regions, ensuring comprehensive coverage. This has been extended to multi-robot coordination in [24], where robots select distinct frontiers to minimize redundant exploration. Alternative negative feedback mechanisms have also been proposed, such as the stigmergic approach in [25], which implements virtual repellents that discourage robots from revisiting recently explored areas. These approaches share conceptual similarities with our grid-based strategy in maintaining spatial memory to guide exploration, though our method leverages centralized grid-based aggregation rather than distributed frontier selection or stigmergic markers.

\section{The Central Placed Foraging Algorithm}
\label{background}

To establish theoretical control for our study, we utilize the canonical foraging algorithm, CPFA~\cite{BeyondPherom2015}. The robot controller is a finite state machine that governs the transitions between departing, searching, surveying, and returning to a central collection zone as described in~\cite{BeyondPherom2015}:
 (see Fig.~\ref{fig_states}):

\begin{figure}[htbp!]
    \centering
    \includegraphics[width=3.2in]{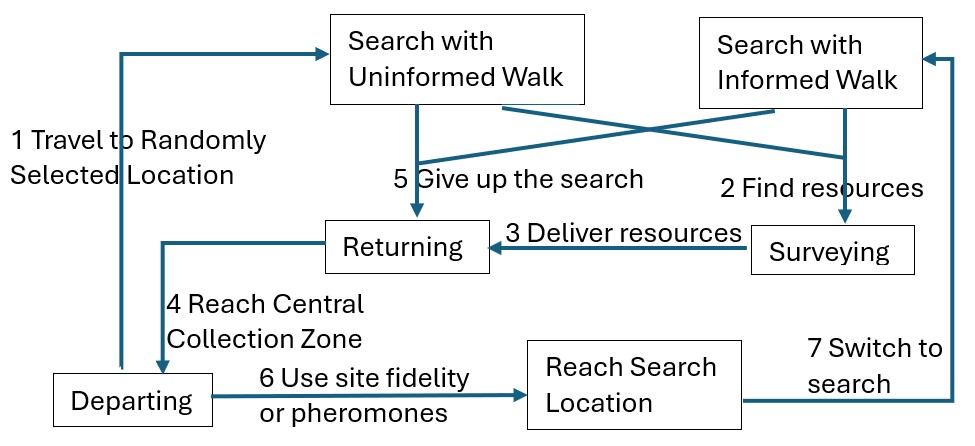}
    \vspace{-1mm}
    \caption{The flow chart of an individual robot's behavior and states in the CPFA}
    \label{fig_states}
\end{figure}

\textbf{Departing}: Robots depart from the central collection zone to a randomly selected location. Robots have the probability to stop on the way at any time and transition to \textit{Searching}. Robots do not have any prior information about the location of resources in the environment. Every robot can remember the location of a previously found resource and can return to this location, a process called \emph{site fidelity} in ant colonies~\cite{Beverly01052009}. The robot cannot remember all previously visited locations and it can only remember the location in the last round of searching. Robots can also communicate using pheromones~\cite{Pheromone2003,Jackson2007}, which are simulated as artificial waypoints~\cite{Campo2010} to recruit robots to known locations with resources. If a robot returns (\textit{Returning}) to the center and then departs, it has the probabilities to depart using \textit{site fidelity} or \textit{pheromone waypoints} in the center. Otherwise, the robot travels to a randomly selected location and repeats the steps as shown in Fig.~\ref{fig_states}.
    
\textbf{Searching}: Robots search for resources using a random walk~\cite{Crist1991}. Successful searches lead to \textit{Surveying}. If a robot found resources, it can only pick up one resource and deliver it to the center. Otherwise, it keeps searching, and it has the probability of giving up the search and returning to the center, which leads to \textit{Returning}. %while unsuccessful ones may end with a return to the center, determined by a probability $p_r$.
    
\textbf{Surveying}: Robots assess the density of local resources within a search radius, recording the number of resources. The density information will be used by the central server in the center to determine the creation of pheromone waypoints~\cite{BeyondPherom2015}. 
    
\textbf{Returning}: Robots deliver resources back to the center. The density of resources $\lambda_{lp}$ is taken into account to generate a probability of laying a new pheromone waypoint. The robots then restart the cycle from \textit{Departing}.

%\subsection{The Baseline: Central Place Foraging Algorithm (CPFA)}
%\label{sec: AtkSetup}

The exploration strategy, ``Uninformed Search," is triggered when a robot has no prior information about the locations of resources (pheromone or site fidelity) and is unable to exploit them. In this state, the robot selects a random location close to the boundaries of the arena. While traveling toward this target location, the robot has a probability (76\% every 5 seconds) of transitioning into a Correlated Random Walk (CRW). This CRW is a stochastic movement pattern in which the robot’s heading is updated by a random angle, and the step size is fixed.
\section{Method}
\label{method}

%We proposed to improve the stochastic/random foraging algorithm CPFA further. 
%In the CPFA algorithm, robots search for resources randomly and share detected resource locations with other robots. Therefore, some resource locations are visited multiple times, while others are never visited within a certain time window. If we expect to collect all resources in a large, unknown environment, robots should visit all resource locations. The time of collecting all resources is significantly higher than collecting resources as much as possible in a certain time window. In~\cite{CompleteCollection2015}, the experimental results indicate that collecting the last $12\%$ of the total resources takes $63\%$ and $75\%$ of the total foraging time. In other words, when the density of resources is low in the arena, the random search is inefficient. 

%In many existing foraging algorithms for robot swarms~\cite{BeyondPherom2015,MatthewDDSA2016,Lu2018DynamicDepots,2021RobotChain}, robots search for and collect resources as much as possible in a certain time window. 
%Although the CPFA excels at exploitation through biologically inspired mechanisms (pheromones and site fidelity), its exploration mechanism is fundamentally stochastic and informed by site fidelity and pheromone waypoints for recruitment. Our proposed algorithm extends this stochastic exploration with an additional grid-based, memory-enhanced strategy designed to maximize area coverage and minimize redundant search operations. 

%\subsection{The Proposed Method}

If we expect to collect all resources, the foraging time will increase significantly in the current uninformed random search strategy~\cite{CompleteCollection2015}. When the number of resources is scarce, e.g., $\le 10\%$ of total resources (or the density of resources is very low), it is challenging to visit all the locations of resources. If robots can remember the locations they visited and share them with the central server when they return to the center, the server can organize the visited locations and guide robots to go to regions where they are least visited. This can improve the possibility of finding resources in the least visited locations. 

However, we cannot let robots remember all the locations they visited, since it requires a lot of memory and computational cost. When the arena size and the number of robots in a swarm are large (e.g., hundreds or thousands), it is infeasible in a real-world environment since the simulated locations can be images, videos, or 3D point clouds. If the space is continuous, it is impossible to record all visited locations. Robots and the central server have limited memory, and the server has limited computation capability in processing information from the robot swarms. When the number of visited locations is large, it takes time to select an unvisited location. Robots need to iterate the entire list of visited locations to identify an unvisited location. If the number of robots is $m$, a robot returns to the center $c$ times at maximum, and the maximum number of visited locations is $r$, the required memory space is $O(mcr)$.

Based on the limitations, we proposed a grid-based visit-count-guided location selection strategy to improve the CPFA search strategy. The design is shown below.

% 1) Memory and Sampling: As shown in the RobotSearchLoop of Algorithm~\ref{alg1}, each robot maintains a local FIFO (First-In-First-Out) queue with a maximum capacity of $20$ locations (it can be a larger number, e.g., 50, based on the arena size). It adds a location to the queue every 25 seconds (estimated based on the robot's memory and the arena size) to capture a general snapshot of its path when searching. When a robot returns to the central collection zone, it offloads this data to the server, clearing its local memory for the next journey. We assume there is no direct communication between robots. Robots only upload visit histories and receive target regions when returning to the central server, which reduces the communication throughput or failure. When robots depart the central server, they make decisions by themselve which is in decentralized mode.
1) Memory and Sampling: As shown in the RobotSearchLoop of Algorithm~\ref{alg1}, each robot maintains a local FIFO (First-In-First-Out) queue with a maximum capacity of $20$ locations. The locations are logged every 25 seconds. 

The two selected numbers were chosen based on our experience. The maximum capacity can be increased to 100 and the frequency to every second. Here, we intentionally aim to demonstrate the limited memory and storage capacity of an individual robot within the swarm. Otherwise, requiring large memory or capacity for each robot would reduce the scalability of the swarm system.

The maximum memory capacity of 20 recent positions similarly reflects this trade-off: the queue must be large enough to describe meaningful spatial coverage during the search phase, yet it has to be small enough and scalable to remain feasible on resource-constrained hardware that we assume is used in swarm robotics.

This interval balances two competing concerns: when we log too frequently, it exhausts onboard memory and inflates communication overhead on each server return; in contrast, logging too infrequently produces a stale visit map that does not give a good representation of the robot movement in the searching phase.  Optimizing these parameters and finding the right balance for different scenarios is left for future work. When a robot returns to the central collection zone, it offloads this data to the server, clearing its local memory for the next journey. We assume that there is no direct communication between robots. Robots only upload visit histories and receive target regions when returning to the central server, which reduces throughput or failure. When robots leave the central server, they make decisions by themselves, which is in decentralized mode.

\begin{algorithm}
\caption{Grid-Based Complete Foraging Algorithm (GCFA)}\label{alg:gcfa}
\begin{algorithmic}[1]
\Procedure{RobotSearchLoop}{}
    \State $Q \gets \emptyset$ \Comment{Initialize visit queue, max size 20}
    \While{searching}
        \State \textbf{wait} 25 seconds
        \State $(x, y) \gets$ current position
        \State $Q.\text{append}((x, y))$
        \If{$|Q| > 20$} $Q.\text{pop\_front}()$ \EndIf
    \EndWhile
    \State \textbf{return to} Central Collection Zone
    \State \Call{UpdateServer}{$Q$}
\EndProcedure

\Procedure{UpdateServer}{$Q$}
    \For{$(x, y) \in Q$}
        \State $cell \gets \text{IdentifyCell}(x, y, n \times n)$
        \State $cell.count \gets cell.count + 1$
    \EndFor
    \If{state = Uninformed Walk}
        \State $R \gets \text{SelectRegion}(\min \sum 3 \times 3 \text{ counts})$
        \State \Call{ExecuteGridSpiral}{$R$}
    \EndIf
\EndProcedure
\end{algorithmic}
\label{alg1}
\end{algorithm}

\begin{figure}[htbp!]
    \centering
    \includegraphics[width=1.8in]{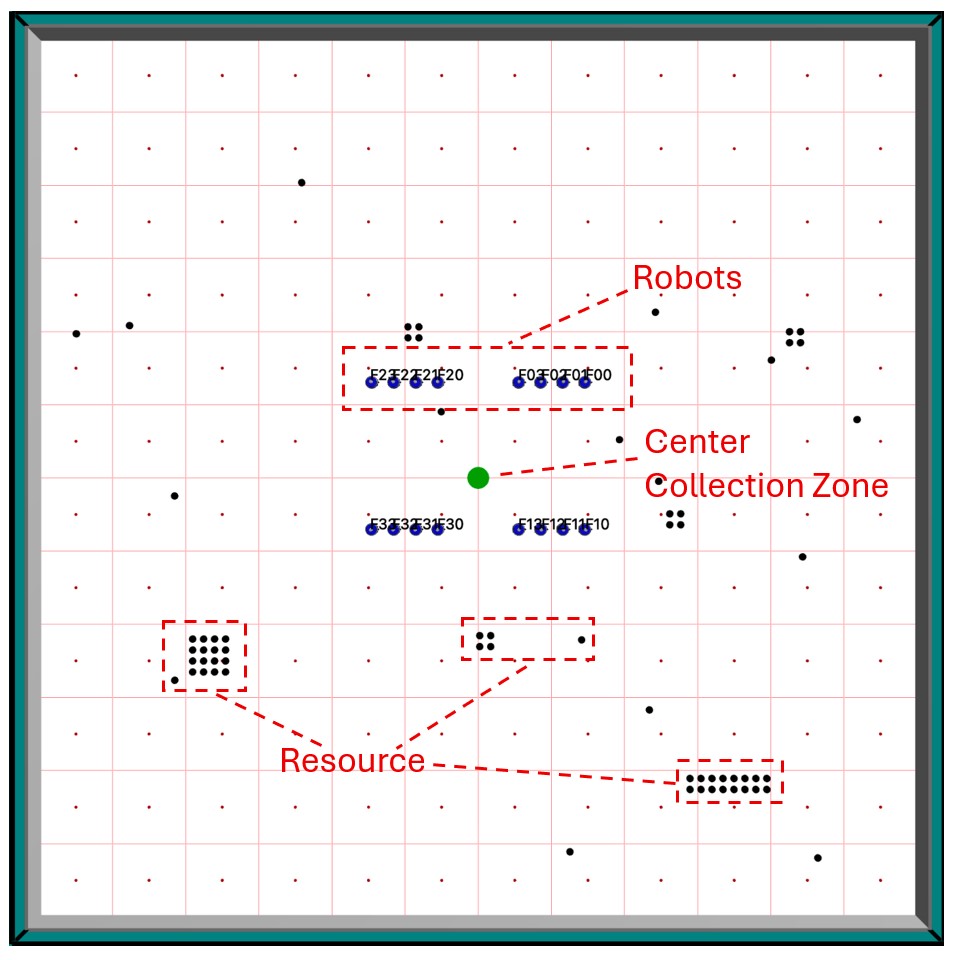}
    \vspace{-2mm}
    % \caption{The grid of a $12 \times 12$ meter arena in ARGoS simulation}
    \caption{ARGoS simulation showing robots, central collection zone, and resources around the arena. Robots and the central zone remain consistent while resources vary by distribution type.}
    \label{fig_grid}
\end{figure}

\begin{figure}[thpb]
\centering
\subfloat[]{
\label{fig_grid_weights}
  \includegraphics[width=0.23\textwidth]{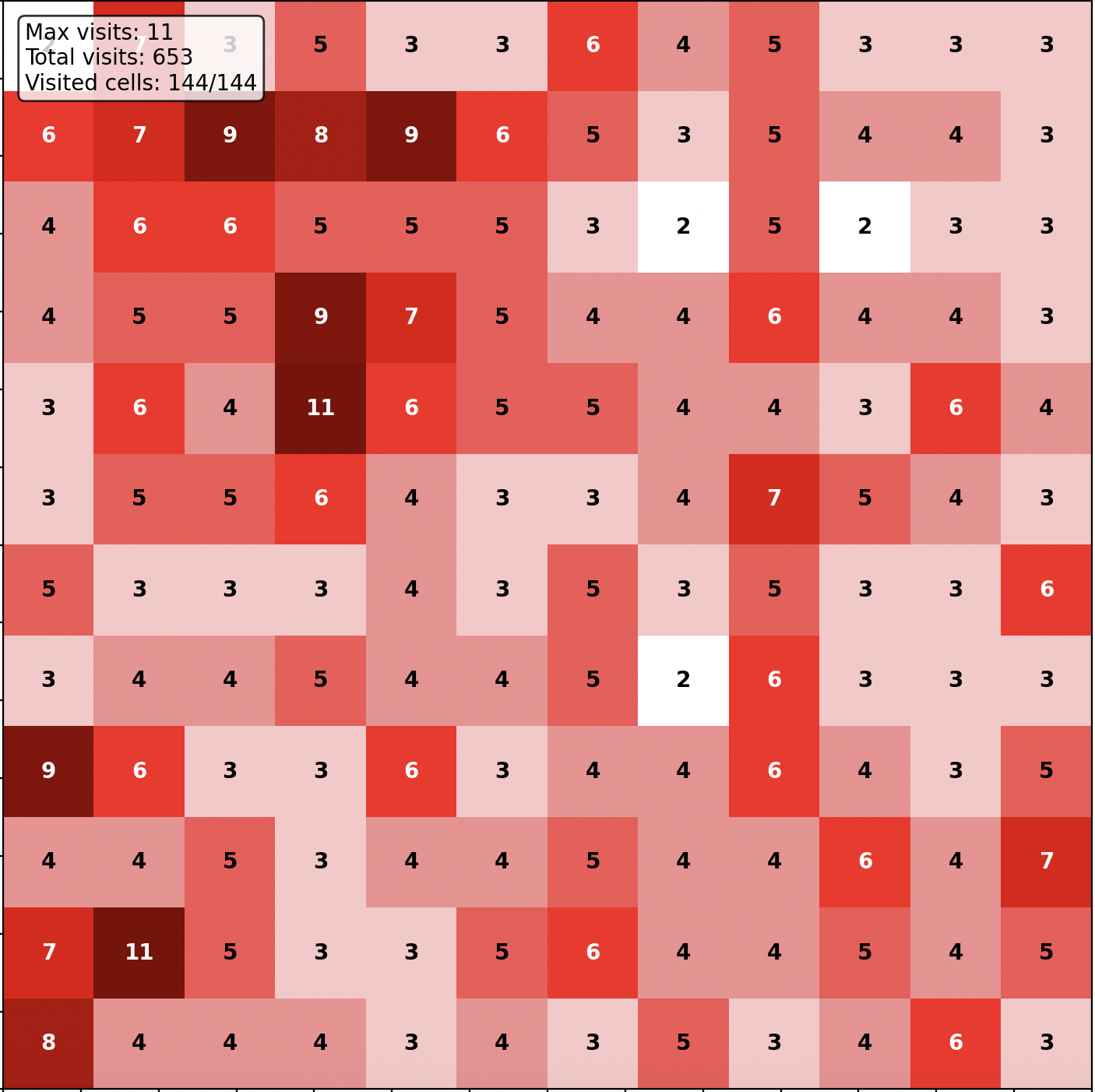}
}
%\vspace{1mm}
\subfloat[]{
\label{fig_grid_spiral}
  \includegraphics[width=0.232\textwidth]{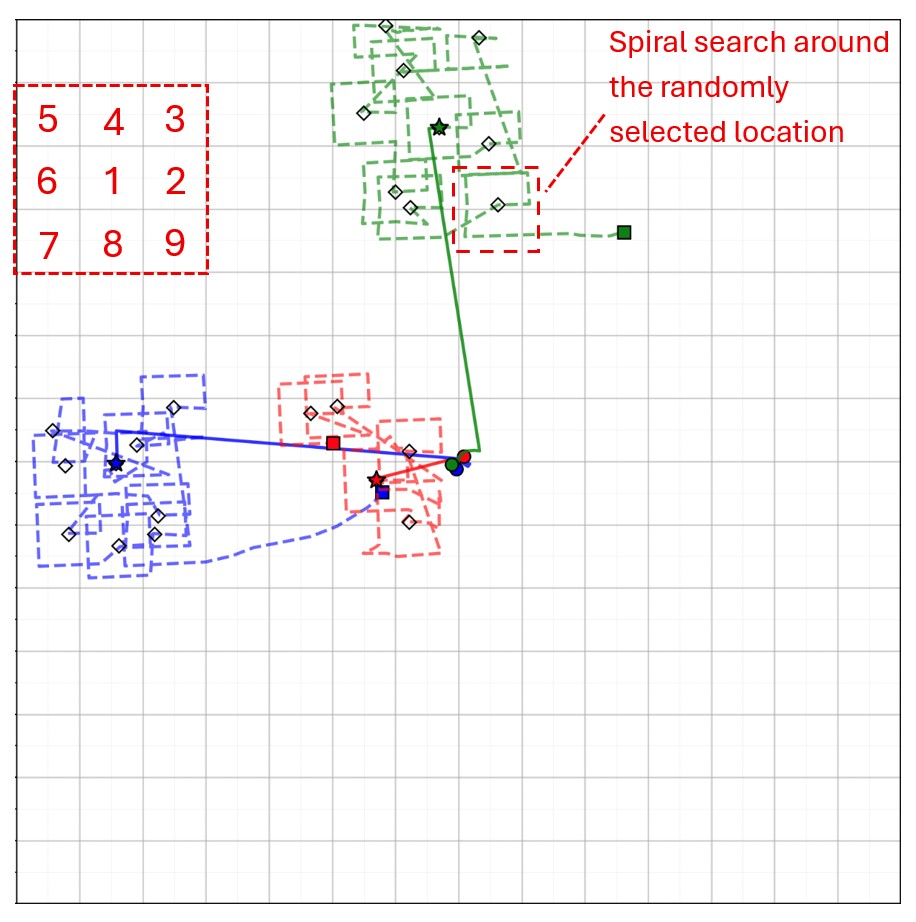}
}
\caption{Illustration of the visitation grid and the spiral search in selected cells. (a) Heatmap of visitation grid for the $12\times12$~m arena. The numbers are the visit counts in cells. Darker red cells have been visited more often, and (b) Counter-clockwise spiral search of the selected $3\times3$ cell block.}
\label{fig_grid_spiral_illustration}
\end{figure}

2) We create a $n\times n$ grid for the search arena based on the size of the arena and the memory capacity of the server (see Fig.~\ref{fig_grid}). We assume that robots have access to global $(x, y)$ coordinates via onboard sensors (e.g., via odometry or GPS receivers) to map visits into the $n \times n$ grid. Robots deliver resources and share visited locations with the central server when they return to the server. Each cell has a counter (weight) (see Fig.~\ref{fig_grid_weights}). If a visited location is in a cell, the counter is increased by one. Therefore, the server does not need to remember all reported locations of the robot swarms. Only the counters need to be updated in the cells, which can be handled in $O(n^2)$ memory space.  

\begin{comment}
    
\begin{figure}[htbp!]
    \centering
    % --- Left Image ---
    \begin{minipage}{0.48\linewidth}
        \centering
        \includegraphics[width=\linewidth]{Figures/grid_weight_12.png}
        \caption{Visit-count heatmap for the $12\times12$~m
                 arena. Darker red cells have been visited
                 more often.} %robots are directed toward the lightest (least-visited) $3\times3$ block.
        \label{fig_grid_weights}
    \end{minipage}
    \hfill
    % --- Right Image ---
    \begin{minipage}{0.49\linewidth}
        \centering
        \includegraphics[width=\linewidth]{Figures/Spiral_Simulation_Updated.jpg}
        \caption{Counter-clockwise spiral search of the
                 selected $3\times3$ cell block. } % Green/blue Cells are fully searched; red cell is in-progress or abandoned (resource found or search given up).
        \label{fig_grid_spiral}
    \end{minipage}
\end{figure}
\end{comment}

3) When a robot returns to the center, the server identifies a block $3\times 3$ of cells with the minimum cumulative visit count.  To ensure complete local coverage, the robot picks a random location in each one and visits these 9 cells in a fixed counter-clockwise search order (see the 9 cells with numbers in Fig.~\ref{fig_grid_spiral}). When the robot reaches a random location in each cell, it runs a spiral search. If a robot finds a resource, it delivers the resource to the center (see the blue spiral search path). If it completes the search of all $9$ cells and has not found a resource (see the green spiral search path), it performs a correlated random walk until a) a resource is found or b) it randomly gives up searching with the probability $p$. Otherwise, it delivers the collected resource to the center. As described in the ExecuteGridSpiral procedure of Algorithm~\ref{alg1}, this systematic approach directly addresses the ``endgame slowdown" by reducing redundant searches in areas already cleared of resources.

In Fig.~\ref{fig_grid_spiral}, the spiral search of $9$ cells is completed in green and blue. The robots then continue the search randomly since they do not find any resources. The one in red does not complete the search for the 9 cells. The robot may give up searching or find a resource. Then, it goes to the center. The experiment videos are available on YouTube\footnote{\url{https://tinyurl.com/bdfdj5p2}}.   

\section{Experimental Setup}
\label{experiment}

The experiments are designed to evaluate the foraging performance of our proposed search strategy to collect all resources under various unknown environments. The experimental parameters are detailed in Table~\ref{tab_combined_parameters}. The configuration settings include various arena sizes, resource distributions, and robot numbers. Our experiments ensure repeatability and reliability.

\begin{table}[!htbp]
\centering
\caption{Experimental Configurations}
\vspace{-2mm}
\label{tab_combined_parameters}
\begin{tabular}{|l|c|c|c|}
\hline
Experiments            & I                                                                     & II                                                                    & III                                                                      \\ \hline
Arena Size    & $14\times 14$   & $14\times 14$   & \begin{tabular}[c]{@{}l@{}}$8\times 8$, $10\times 10$, \\ $12\times 12$, $14\times 14$ \\ $16\times 16$ \end{tabular} \\ \hline
\# of Resources        &  \begin{tabular}[c]{@{}l@{}} 16, 32 \\ 48, 64, 80 \end{tabular}   & 48      & 48  \\ \hline
\# of Robots           & 16                                                                    & 16                                                                    & 16                                                                       \\ \hline
\# of Runs             & 30                                                                    & 30                                                                    & 30                                                                       \\ \hline
 \begin{tabular}[c]{@{}l@{}}Resource \\ Distributions\end{tabular} & \begin{tabular}[c]{@{}l@{}}Clustered\\ Powerlaw\\ Random\end{tabular} & \begin{tabular}[c]{@{}l@{}}Clustered\\ Powerlaw\\ Random\end{tabular} & Random                                                                   \\ \hline
\end{tabular}
\end{table}

We conducted three experiments. In the first experiment (Experiment I), we evaluate foraging performance when we vary the number of resources. The resources are in three distributions (Clustered, Powerlaw, and Random)~\cite{BeyondPherom2015}. The resources are clustered in the same size, $8\times 8$, in the Cluster Distribution, $8\times 8$, $4\times 4$, $2\times 2$, and $1$ in the Powerlaw Distribution, and scattered uniformly in the Random Distribution. 
%\textbf{Experiment 1 - Pheromone Trail Dynamics: }

In Experiment II, we evaluate the foraging performance when robot swarms collect every $10\%$ of the resources in three resource distributions. The total number of resources is $48$. In Experiment III, we evaluate the foraging performance when we vary the size of the arena. 

The foraging performance is evaluated by the time to collect all resources. The shorter time indicates a better performance. We compare our proposed algorithm with the original CPFA algorithm.

\section{Results}
\label{results}
The analysis is organized into three key areas: first, we assess the flexibility of the algorithm as the resources increase; second, we examine the algorithm's robustness in different arena sizes; and finally, we conduct an analysis of the final collection phase to quantify the efficiency improvements.  We replicate each experiment in 50 trials and report the median time for the swarm to collect resources in each experiment. We present our results in box plots to show which results are statistically different. The statistical significance is explicitly indicated by asterisks in the figures ($p < 0.001$). 

\subsection{Flexibility with Different Number of Resources}

\begin{figure}[htbp!]
    \centering
    \includegraphics[width=3.3in]{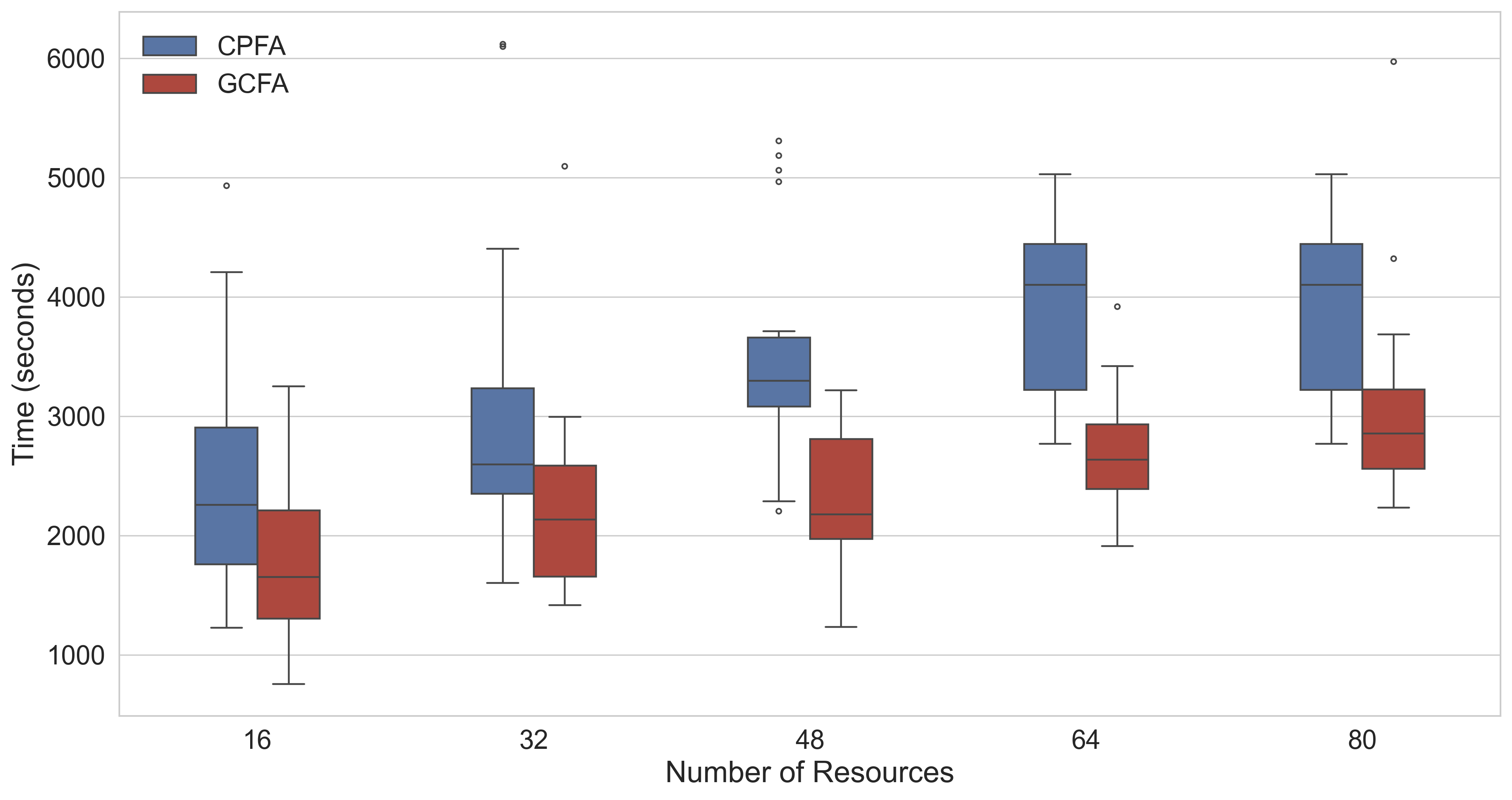}
    \vspace{-2mm}
    \caption{Collection time vs.\ number of resources
             (Random distribution). GCFA consistently
             outperforms CPFA, with the gap widening from
             $\sim$400~s at $N=16$ to $>$800~s at $N=80$.}
    \label{fig_1_random}
\end{figure}

\begin{figure}[htbp!]
    \centering
    \includegraphics[width=3.3in]{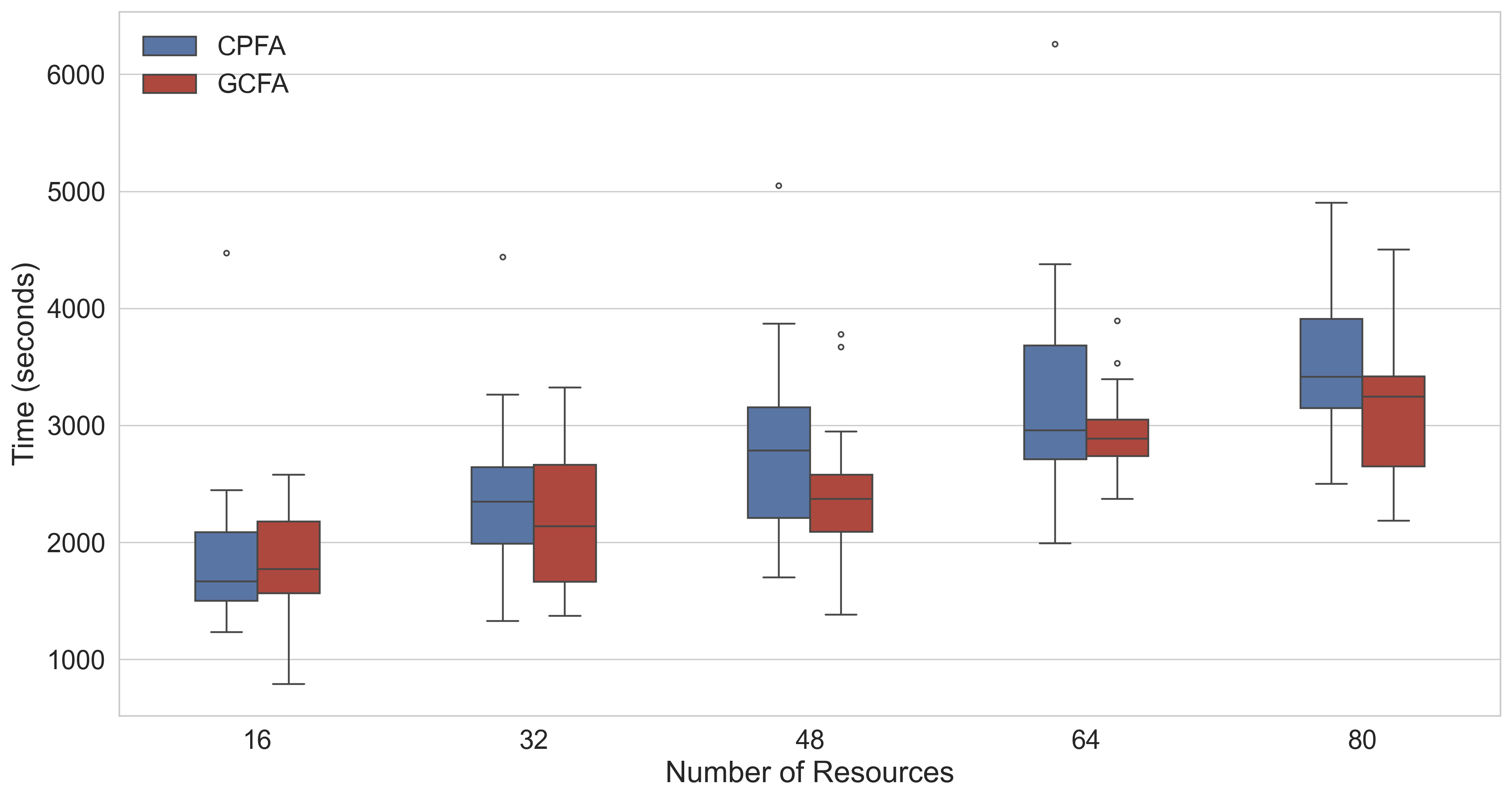}
    \vspace{-2mm}
    % \caption{The time to complete the collection in Powerlaw distribution}
    \caption{Collection time vs.\ number of resources
             (Powerlaw distribution). GCFA advantage grows
             from $\sim$90~s at $N=16$ to $>$360~s at
             $N=80$, demonstrating scalability with resource
             count.}
    \label{fig_1_powerlaw}
\end{figure}

\begin{figure}[htbp!]
    \centering
    \includegraphics[width=3.3in]{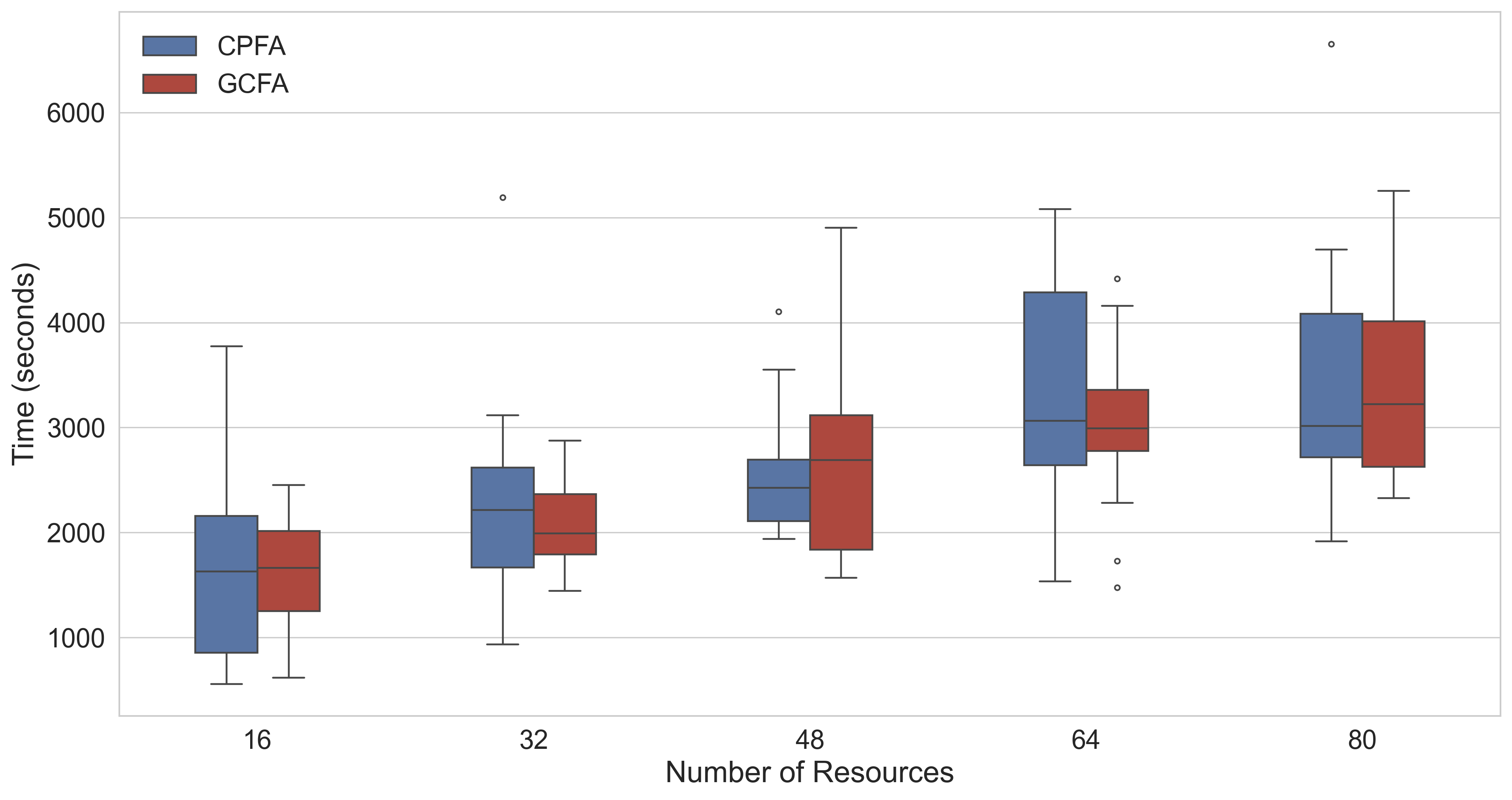}
    \vspace{-2mm}
    % \caption{The time to complete collection in Clustered distribution}
    \caption{Collection time vs.\ number of resources
             (Clustered distribution). No
             significant difference between GCFA and CPFA,
             confirming that visit-count guidance provides
             no benefit when resources are spatially
             concentrated.}
    \label{fig_1_cluster}
\end{figure}

Figs.~\ref{fig_1_random} -~\ref{fig_1_cluster} visualize box plots of different resource distributions with various numbers of resources. In exploration-heavy tasks (Random and Powerlaw distributions), the performance gap between the algorithms widened as the number of resources increased. In the Random distribution, the performance at $N=16$ was about $400$ seconds. At $N=80$, this gap increased to more than $800$ seconds. Similarly, in the Powerlaw distribution, the performance gap at $N=16$ was approximately $90$ seconds. At $N=80$, this gap increased to more than $360$ seconds. This shows that the CPFA algorithm takes more time in the foraging task. The GCFA algorithm maintains a more linear relationship between resource count and collection time, validating its suitability for higher volume foraging tasks. Furthermore, we observed that our GCFA achieved an overall improvement of $33\%$ in the Random distribution and $14\%$ in the Powerlaw distributions compared to the CPFA ($48$ resources). However, there is no difference in the more exploitation-heavy tasks (Clustered distribution).  

\vspace{-3mm}

\begin{comment}
    
\begin{figure}[htbp!]
    \centering

    % ---------- Top Row (Two Side-by-Side) ----------
    \begin{minipage}{0.48\linewidth}
        \centering
        \includegraphics[width=\linewidth]{Figures/random_box_plot.png}
        \caption*{(a) Random Distribution}
        \label{fig_1_random}
    \end{minipage}
    \hfill
    \begin{minipage}{0.48\linewidth}
        \centering
        \includegraphics[width=\linewidth]{Figures/powerlaw_box_plot.png}
        \caption*{(b) Powerlaw Distribution}
        \label{fig_1_powerlaw}
    \end{minipage}

    \vspace{4mm}

    % ---------- Bottom Row (One Centered Figure) ----------
    \begin{minipage}{0.55\linewidth}
        \centering
        \includegraphics[width=\linewidth]{Figures/cluster_box_plot.png}
        \caption*{(c) Cluster Distribution}
        \label{fig_1_cluster}
    \end{minipage}

    % ---------- Main caption ----------
    \caption{Completion time distributions across three resource configurations.}
    \label{fig:boxplot_comparison}
\end{figure}

\end{comment}

\subsection{Robustness in Different Arenas}

\begin{figure}[htbp!]
    \centering
    \includegraphics[width=3.5in]{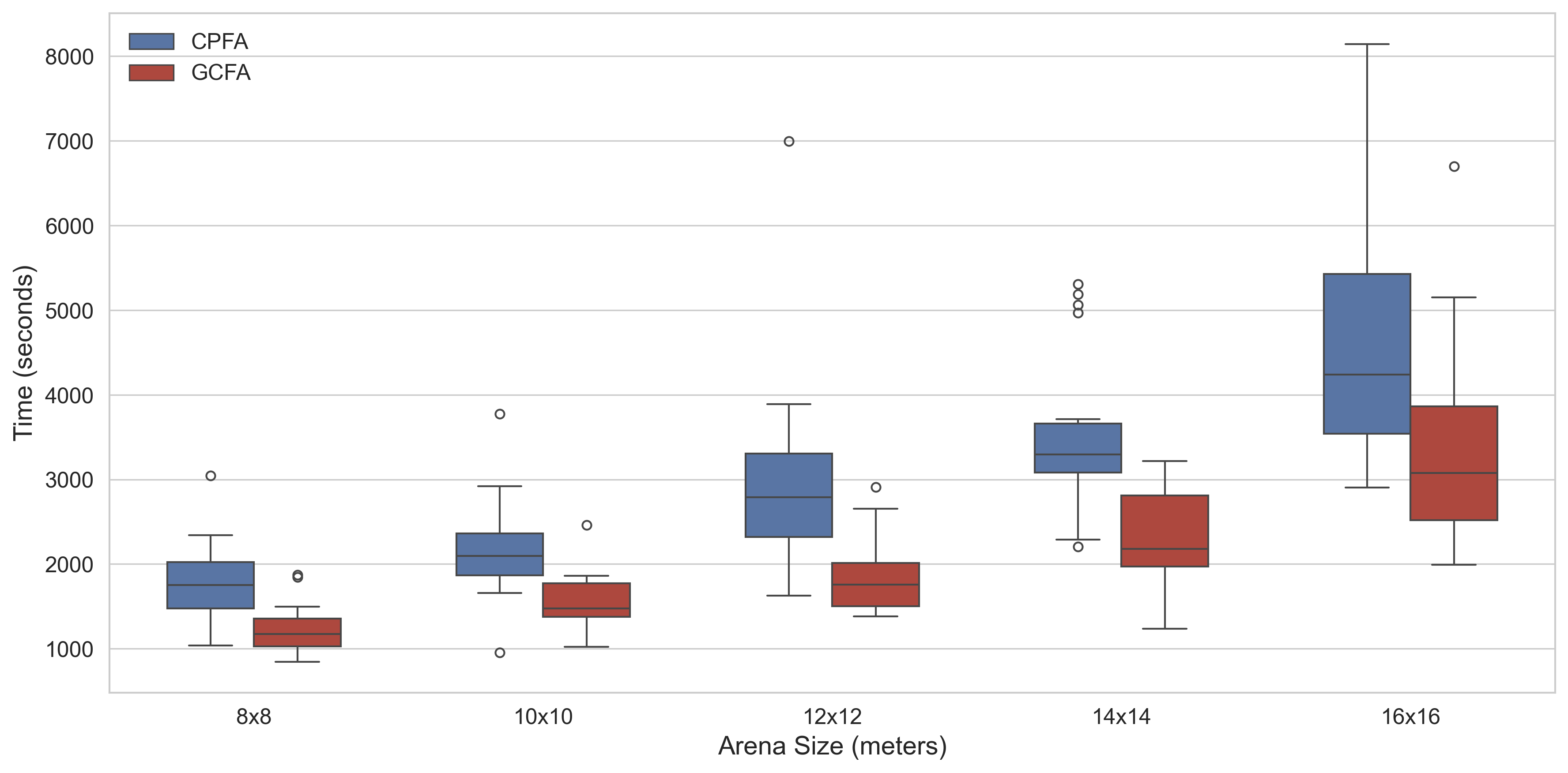}
    \vspace{-2mm}
    % \caption{The time to complete the collection in different arena sizes}
    \caption{Collection time vs.\ arena size (Random dist.,
             $N=48$). The GCFA advantage more than doubles
             from 581~s ($8\times8$~m) to 1{,}267~s
             ($16\times16$~m), showing that the benefit
             scales with the search area.}
    \label{fig_3}
\end{figure}

Robot swarms must work well with foraging algorithms in different arena sizes. Fig.~\ref{fig_3} shows the results with different arena sizes in Experiment III. Foraging performance decreases when the arena size increases because robots travel further. The performance between the two algorithms grows significantly as the arena size increases (see Table \ref{tab_arena_scalability_clean}).

%\vspace{-3mm}

%\vspace{-3mm}

\textbf{Small vs. Large Arenas:}
In the smallest arena ($8 \times 8$ m), the density of the robots is highly relative to the arena size. Even with random movements, agents cover the map quickly, resulting in a moderate performance gap of $581$ seconds. However, as the size of the arena quadruples to $16 \times 16$ m, the performance of CPFA severely degrades. The completion time jumps to $4,564$ seconds for CPFA, compared to $3,297$ seconds for GCFA.

\begin{table}[!htbp]
\centering
\caption{Impact of Arena Size on Completion Time (Random Dist., $N=48$)}
\vspace{-2mm}
\label{tab_arena_scalability_clean}
\begin{tabular}{|l|c|c|c|}
\hline
Arena Size (m)
& GCFA 
& CPFA 
& Time Saved 
\\ \hline
$8 \times 8$   
& 1222 & 1803 & \textbf{581} 
\\ \hline
$10 \times 10$ 
& 1541 & 2165 & \textbf{624} 
\\ \hline
$12 \times 12$ 
& 1827 & 2973 & \textbf{1146} 
\\ \hline
$14 \times 14$ 
& 2355 & 3540 & \textbf{1184} 
\\ \hline
$16 \times 16$ 
& 3298 & 4564 & \textbf{1267} 
\\ \hline
\end{tabular}
\end{table}

GCFA saves more than double the time from the smallest to the largest arena ($580 \rightarrow 1,266$ s). This indicates that systematic exploration becomes increasingly vital as the ratio of search area to agent count increases. In large, sparse environments, the GCFA algorithm prevents robots from re-visiting cleared regions and guides robots to go to unvisited regions, providing a highly efficient solution for large foraging environments.

%\vspace{-5mm}
% SECTION 3: LATE GAME (THE EXPLANATION)
% Now we explain WHY the scalability sections looked that way.

\subsection{Analysis of the Final Collection Phase}

To identify what drives the performance differences observed in the flexibility experiments, we analyzed the timing of resource collection. Autonomous foraging is rarely linear; swarms typically slow down significantly near the end of a mission. As resource density approaches zero, the time required to locate the final few items often increases exponentially for memoryless systems, like CPFA.

\begin{figure}[htbp!]
    \centering
    \includegraphics[width=3.3in]{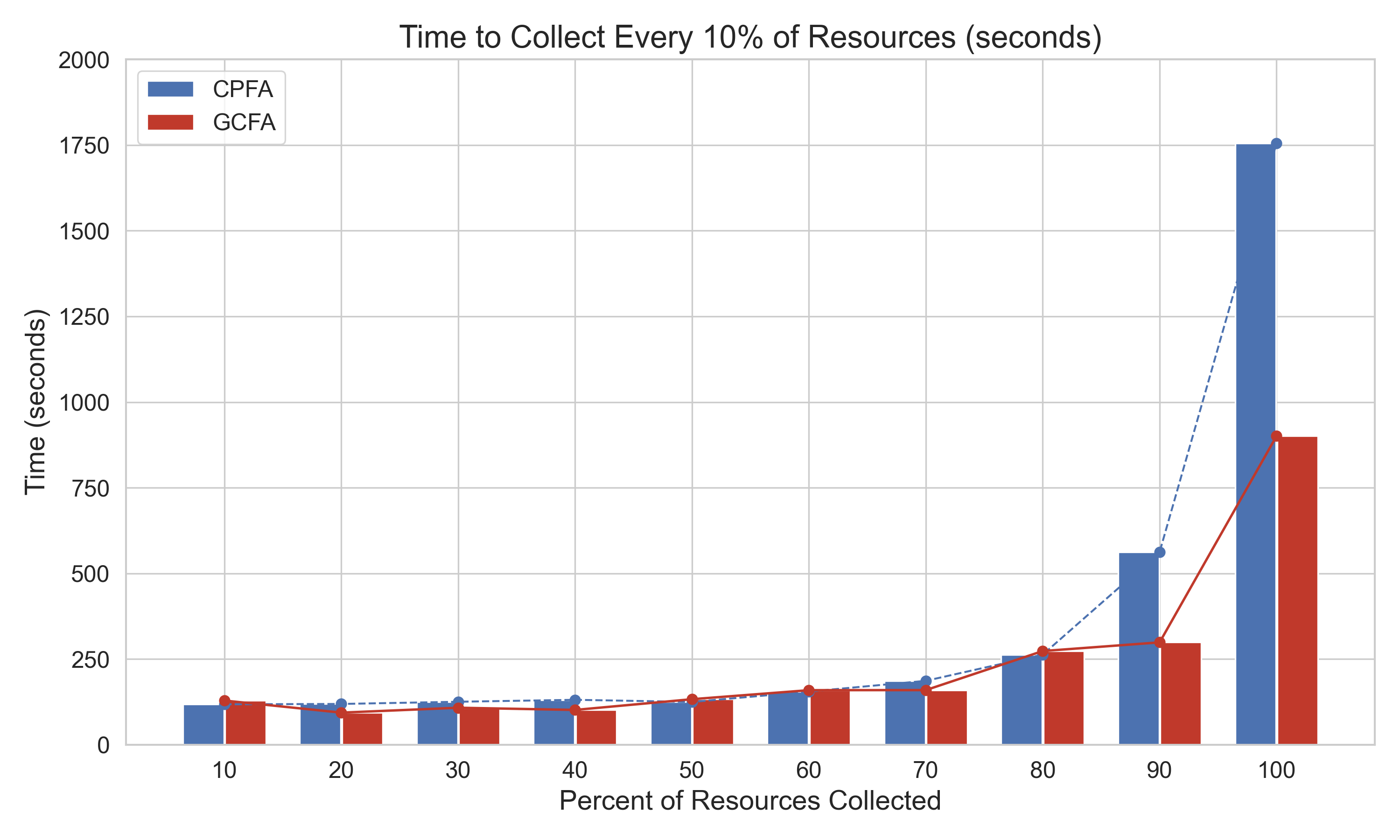}
    \vspace{-2mm}
    % \caption{The time to collect every $10\%$ of resources in Random distribution}
    \caption{Time to collect each successive 10\% of
             resources (Random dist., $N=48$). CPFA time
             triples in the final interval
             ($563\to1{,}755$~s); GCFA reduces this to
             901~s --- a 49\% improvement.}
    \label{fig3_Random}
\end{figure}

\begin{figure}[htbp!]
    \centering
    \includegraphics[width=3.3in]{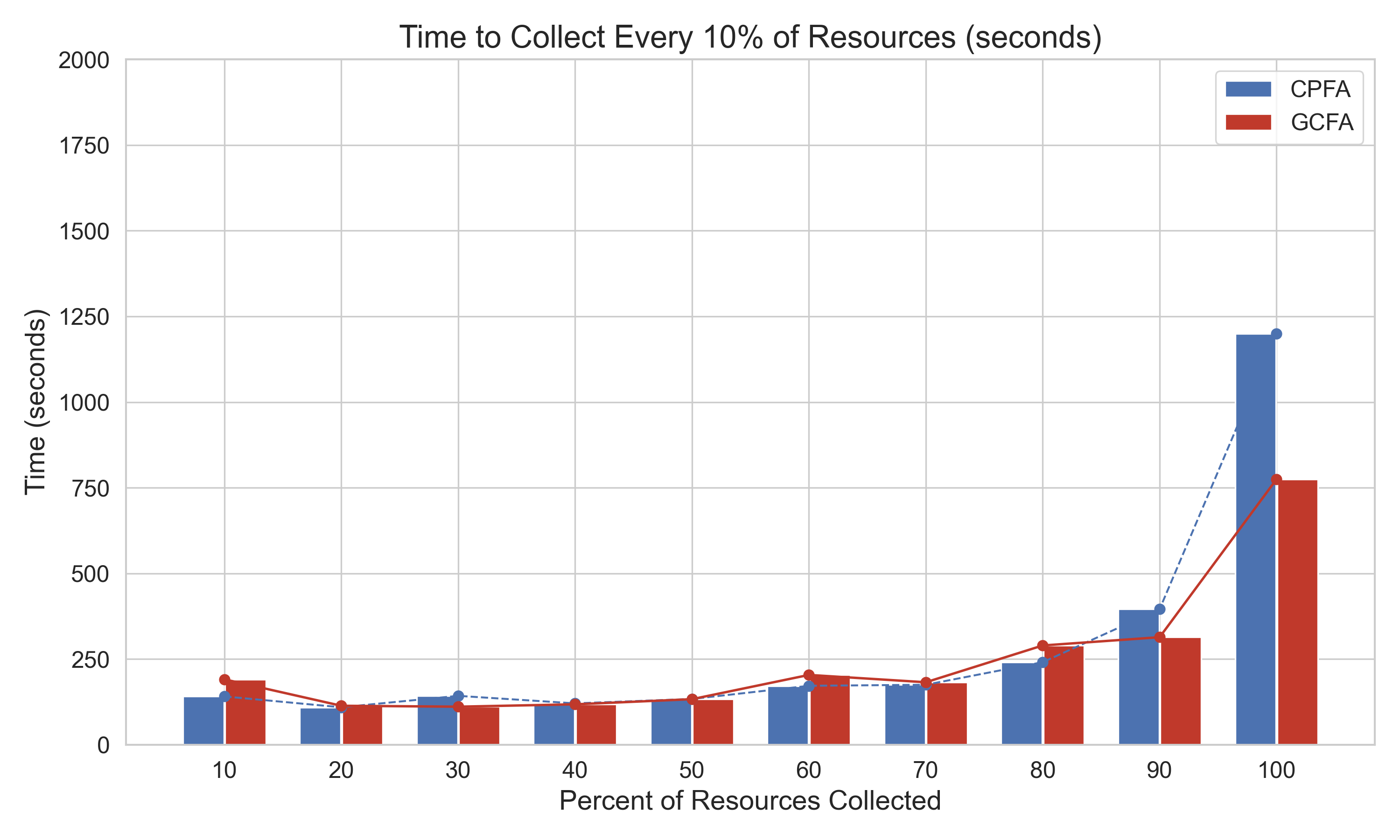}
    \vspace{-2mm}
    % \caption{The time to collect every $10\%$ of resources in Powerlaw distribution}
    \caption{Time to collect each successive 10\% of
             resources (Powerlaw dist., $N=48$). GCFA
             reduces final-interval time from 1{,}199~s
             (CPFA) to 775~s --- a 35\% improvement,
             confirming robustness to mixed distributions.}
    \label{fig3_Powerlaw}
\end{figure}

\begin{figure}[htbp!]
    \centering
    \includegraphics[width=3.3in]{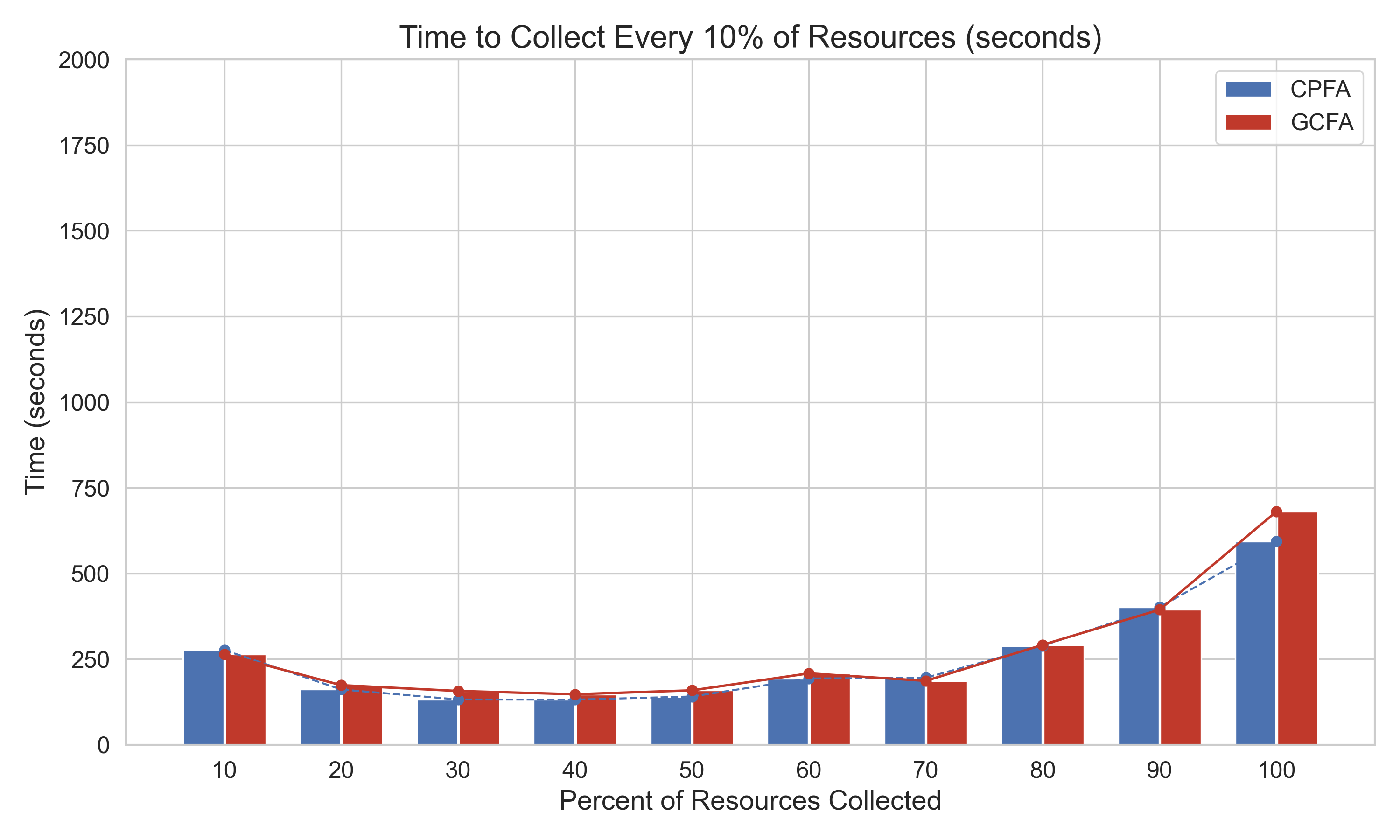}
    \vspace{-2mm}
    % \caption{The time to collect every $10\%$ of resources in Cluster distribution}
    \caption{Time to collect each successive 10\% of
             resources (Clustered dist., $N=48$). CPFA
             outperforms GCFA in the final interval (594~s
             vs.\ 681~s) because remaining resources are
             co-located with previously found clusters.}
    \label{fig3_Cluster}
\end{figure}

We measured two specific phases to see exactly how much difficulty increased: \textbf{Interval A (80\% $\rightarrow$ 90\%):} The phase where resources become scarce, forcing robots to spend more time searching. \textbf{Interval B ($90\% \rightarrow 100\%$):} The final stage, which is the most difficult because the last few items are scattered and hard to find. Figs.~\ref{fig3_Random} -~\ref{fig3_Cluster} show the time to collect every $10\%$ of resources for three resource distributions in Experiment II.
Table~\ref{tab_late_game_clean} details the duration required to complete these intervals. The results show that the biggest strength of GCFA is preventing this slowdown at the end of the mission.

%\vspace{-3mm}

\begin{table}[!htbp]
\centering
\caption{Time of the Final Collection Phases}
\vspace{-2mm}
\label{tab_late_game_clean}
\resizebox{1.0\columnwidth}{!}{
\begin{tabular}{|c|c|c|c|c|c|}
\hline
Config.  & \begin{tabular}[c]{@{}l@{}}CPFA\\ $80\% \rightarrow 90\%$ \end{tabular} & \begin{tabular}[c]{@{}l@{}}GCFA\\ $80\% \rightarrow 90\%$ \end{tabular} & \begin{tabular}[c]{@{}l@{}}CPFA\\ $90\% \rightarrow 100\%$ \end{tabular} & \begin{tabular}[c]{@{}l@{}}GCFA\\ $90\% \rightarrow 100\%$ \end{tabular} & \begin{tabular}[c]{@{}l@{}}Endgame\\ Improvement\end{tabular} \\ \hline

Random   & 562.5     & 298.6      & 1755.2       & 900.8    & \textbf{+48.68\%}   \\ \hline
Powerlaw & 395.5     & 313.7      & 1199.2       & 774.8    & \textbf{+35.39\%}   \\ \hline
Cluster  & 401.4     & 394.0      & 593.5        & 680.7    & -14.69\%            \\ \hline
\end{tabular}
}
\end{table}

%\subsubsection{Random Distribution Analysis}
 In the random distribution, the CPFA algorithm suffered a severe performance penalty in the final phase (see Fig.~\ref{fig3_Random}). Although CPFA performed adequately in the early stages, the duration increased from $563$ seconds (Interval A) to $1,755$ seconds (Interval B). This roughly $3\times$ more difficulty indicates that correlated random walks are inefficient when the resource density is low, and robots spend the majority of their time re-visiting space.

The GCFA algorithm successfully reduced this delay. By retaining a memory of visited cells, the swarm effectively reduced the search space over time and was able to search in a "smart" manner. Consequently, GCFA completed the final interval in $901$ seconds —nearly twice as fast as CPFA.

%\subsubsection{Power Law Distribution Analysis}\

In the powerlaw distribution, dense clusters are combined with scattered resources. The data mirrors the Random distribution trend but highlights the GCFA's robustness. The search time of CPFA jumped to $1,199$ s in the final interval (see Fig.~\ref{fig3_Powerlaw}). The GCFA algorithm proved far more resilient, completing the final interval in 775 seconds, a 35\% reduction.

%\subsubsection{Cluster Distribution Analysis}
In the clustered distribution, the results offer a different insight that validates the trade-off between exploration and exploitation. In this scenario, the GCFA algorithm was outperformed by the CPFA, specifically in the final interval (CPFA: $594$ s vs. GCFA: $681$ s) (see Fig.~\ref{fig3_Cluster}). Because the final resources were located adjacent to or very close to previous finds (within dense clusters), the CPFA's tendency to stay and search in one spot became an advantage. In contrast, the GCFA algorithm's strategy to check unvisited areas backfired, as it forced robots to travel to empty corners of the arena rather than persisting at known cluster sites.

\subsection{Efficiency of GCFA}
 
The results above demonstrate consistent time savings that, while not order-of-magnitude, translate into concrete operational advantages. We address two main aspects of the practical case for choosing the proposed GCFA.
 
\textit{Complete collection is a time-minimization problem:} In real-world applications --- search and rescue, humanitarian demining, chemical-leak detection --- the mission cannot be declared complete until all (or nearly all) targets have been retrieved. Both GCFA and CPFA eventually achieve $100\%$ collection, and the search is stochastic, so the \emph{only} dimension on which they can differ is the time required to reach that goal. A $33\%$ reduction in total collection time in Random distribution, and a $48\%$ reduction in the critical final phase (Table~\ref{tab_late_game_clean}), directly translates to shorter robot exposure, lower energy consumption, and faster redeployment.
  
\textit{GCFA is a zero-cost drop-in enhancement:} GCFA does not require additional hardware, no direct robot-to-robot communication, and negligible onboard memory beyond a bounded queue of at most $Q_{\max}$ positions. For any team already running a CPFA algorithm in a Random or Powerlaw resource environment, the adoption cost is effectively zero, while the expected benefit is a consistent reduction in mission time. Therefore, GCFA is straightforwardly preferable.

\section{Conclusion}
\label{conclusion}

% This study proposed a Grid-Based Complete Resource Foraging Algorithm (GCFA) that is designed to address the challenge of complete resource collection in robot swarms. With a shared grid-based map of the arena in the central server, robots select unvisited or less-visited regions rather than random searching areas.
This study proposed the GCFA foraging algorithm, a \emph{visitation grid guided search strategy} to address the challenge of complete resource collection in robot swarms. The key contribution is a lightweight spatial memory mechanism: robots periodically record their positions in a bounded local queue and offload this data to a central server, which aggregates visit counts in a compact $n\times n$ grid. Robots leaving in uninformed search mode are then probabilistically directed toward the least-visited $3\times 3$ neighborhood, reducing redundant revisits without requiring direct robot-to-robot communication or unbounded onboard storage. We also consider the scalability of the robot swarms by limiting robots' and the server's memories, communication throughput, and computation cost.

The results indicate that our method significantly outperforms the canonical CPFA foraging algorithm in scenarios with high exploration requirements. GCFA achieved significant improvements in random resource distributions and power-law distributions. Furthermore, the algorithm can alleviate exponential growth in time in the last ~$20\%$ of resources, increasing efficiency by over $48\%$ during this final collection phase. Robustness evaluation also confirmed that the benefits of our visit-count-guided approach increase with arena size.

However, CPFA outperformed, or had very similar results to, our GCFA method in clustered environments. This highlights that while visit-count guided map exploration is superior for sparse or scattered resources, a balance between exploration and local exploitation remains critical for highly clustered distributions. Cells near resource clusters accumulate high visit counts, reducing their likelihood of being selected during an uninformed random search. As a result, any remaining resources in these clusters become difficult to discover if they were not found earlier through pheromone waypoints.

The running time to select a region for an uninformed search is $O(n^2)$, where $n$ is the number of cells. Specifically, sorting cells by visit count requires $O(n^2)$ time, which dominates the overall computation. The number of cells, therefore, affects both the foraging performance and the running time. A larger $n$ results in smaller cell sizes and a greater number of unvisited cells, which can improve the foraging efficiency. However, increasing $n$ also leads to a higher computational cost. The findings will benefit other stochastic search strategies for robots with limited resources and capabilities. In future work, we will conduct a theoretical analysis of the relationship between the number of cells and the computation cost and investigate strategies to optimize cell selection.


%%%%%%%%%%%%%%%%%%%%%%%%%%%%%%%%%%%%%%%%%%%%%%%%%%%%%%%%%%%%%%%%%%%%%%%%%%%%%%%%

%%%%%%%%%%%%%%%%%%%%%%%%%%%%%%%%%%%%%%%%%%%%%%%%%%%%%%%%%%%%%%%%%%%%%%%%%%%%%%%%

%%%%%%%%%%%%%%%%%%%%%%%%%%%%%%%%%%%%%%%%%%%%%%%%%%%%%%%%%%%%%%%%%%%%%%%%%%%%%%%%
%\section*{APPENDIX}


\section*{ACKNOWLEDGMENT}

The authors acknowledge the financial support provided by the NSF Expand AI program (No. 2434916), the NSF CREST Center for Multidisciplinary Research Excellence in Cyber-Physical Infrastructure Systems (MECIS) (No. 2112650), and the NSF MSI program (No. 2318682).

%This work is supported by the SLA (Scientific Leadership Award) program through DHS (Department of Homeland Security) Award No. 21STSLA00009-01-00. The authors also acknowledge the partial funding provided by the CREST MECIS program through NSF (National Science Foundation) Award No. 2112650 and the MSI program through NSF Award No. 2318682.

%%%%%%%%%%%%%%%%%%%%%%%%%%%%%%%%%%%%%%%%%%%%%%%%%%%%%%%%%%%%%%%%%%%%%%%%%%%%%%%%

%References are important to the reader; therefore, each citation must be complete and correct. If at all possible, references should be commonly available publications.

\bibliographystyle{IEEEtran}
\bibliography{BibFiles/IEEEabrv, BibFiles/main_ref}

\end{document}